\documentclass[11pt]{article}

\usepackage[preprint]{acl}

\usepackage{times}
\usepackage{latexsym}
\usepackage[T1]{fontenc}
\usepackage[utf8]{inputenc}
\usepackage{microtype}
\usepackage{inconsolata}
\usepackage{graphicx}
\usepackage{booktabs}
\usepackage{multirow}
\usepackage{amsmath}
\usepackage{amssymb}
\usepackage{xcolor}
\usepackage{xspace}
\usepackage{enumitem}
\usepackage{colortbl}
\usepackage{array}
\usepackage{tcolorbox}

\usepackage[T1]{fontenc}

\usepackage[utf8]{inputenc}

\usepackage{microtype}

\usepackage{inconsolata}

\usepackage{graphicx}

\definecolor{rowgray}{HTML}{F2F2F2}
\definecolor{navy}{HTML}{084A91}
\definecolor{inkgray}{HTML}{707070}

\newcommand{\swsp}{\mbox{Self-WSP}\xspace}
\newcommand{\qwen}{Qwen3.5-0.8B\xspace}
\newcommand{\code}[1]{\texttt{\small #1}}
\newcommand{\alog}{\code{A\_log}\xspace}
\newcommand{\ns}{\rlap{\smash{\textsuperscript{\dag}}}}
\newcommand{\cp}{\rlap{\smash{\textsuperscript{$\ast$}}}}
\definecolor{grpblue}{HTML}{EAF1FA}
\newcommand{\grpn}[2]{\rowcolor{grpblue}\multicolumn{#1}{@{\hspace{3pt}}l}{\color{navy}\textbf{#2}} \\}
\newcommand{\grp}[1]{\addlinespace[2pt]\grpn{6}{#1}}
\newcommand{\mdl}[2]{\addlinespace[2.5pt]\multicolumn{6}{@{}l}{\textbf{#1}\quad{\color{inkgray}#2}} \\[0.6pt]}

\setlist[itemize]{leftmargin=1.1em,itemsep=1pt,topsep=2pt,parsep=0pt}

\title{Self-WSP: Can Language Models Improve Themselves Through Weight-Space Programming?}

\author{Guoming Ling\textsuperscript{1,2},\quad Shanshan Zhong\textsuperscript{1},\quad Kaixuan Zhang\textsuperscript{2},\quad Zhongzhan Huang\textsuperscript{2} \\
  \textsuperscript{1}Carnegie Mellon University \quad \textsuperscript{2}MilkThink
}

\begin{document}
\maketitle

\begin{abstract}
Frontier language models can now read and edit the weights of a neural network much like code and thereby improve a small model, an approach known as Weight-Space Programming (WSP). We further explore applying WSP to a language model itself, which we call \swsp. Specifically, \qwen reads its own weights, proposes an edit, and then continues with the edited weights as both problem solver and editor. We find many cases of genuine self-improvement. Self-edits can raise GSM8K accuracy from 11.8\% to 62.2\% and MATH500 accuracy from 7.0\% to 32.2\%, sometimes by changing only 16 parameters, and in some cases repeated self-editing keeps improving the model. A closer look, however, reveals that these gains lack interpretability. The model shows little understanding of the edits it makes, and WSP driven by random search sometimes performs comparably. \swsp thus holds promise as a path toward recursive self-improvement (RSI), but realizing this promise requires further exploration.
\end{abstract}

\section{Introduction}

A system that improves its own intelligence, and with it its ability to improve itself further, is a long-standing idea known as recursive self-improvement (RSI) \citep{good1966speculations,schmidhuber2007godel}. Language models already improve themselves in several ways. They train on rationales, data or rewards they generate themselves \citep{zelikman2022star,huang2023selfimprove,yuan2024selfrewarding}, rewrite the code that calls them \citep{zelikman2024stop,zhang2025dgm}, or learn to write their own finetuning data \citep{zweiger2025seal}. In each case, however, the model changes its weights only indirectly, through data, code or training.

\begin{figure}[t]
\centering
\includegraphics[width=\columnwidth]{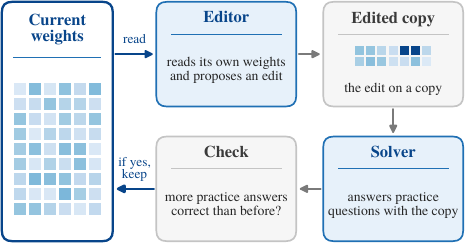}
\caption{\textbf{The \swsp loop.} One model plays two roles. As the editor, it reads its own current weights and proposes an edit. As the solver, it answers practice questions using a copy of the weights with the edit applied. The outer loop checks whether more answers are correct, and if so the copy replaces the current weights, so that the next proposal comes from the edited model.}
\label{fig:loop}
\end{figure}

\begin{figure*}[t]
\centering
\includegraphics[width=\textwidth]{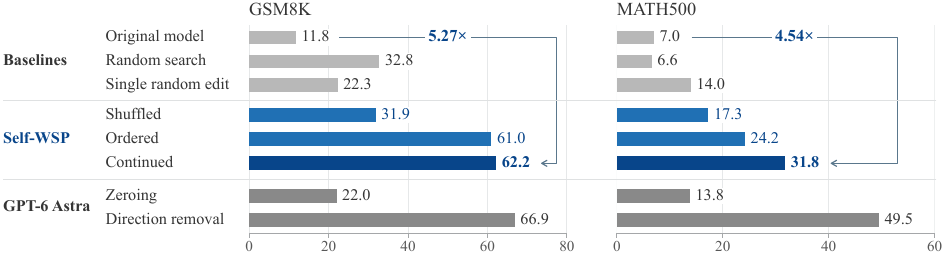}
\vspace{-18pt}
\caption{\textbf{Test accuracy of the best model of each method.} Each method contributes its Top-1 model, the one with the largest gain on validation questions. In the \swsp experiments of Section~\ref{sec:method}, the shuffled interface lists the tensor catalog in random order and the ordered interface lists it by layer. Continued runs let the best shuffled models keep editing themselves under the ordered interface. Bold values mark the highest \swsp score shown for each dataset. Random search replaces the editor with a random sampler under the same budget as the shuffled interface. The single random edit is the best of the edits tested individually on the original model, as described in Section~\ref{sec:random}. GPT-6's two edits come from \citet{takehi2026weightspace} and are re-evaluated in our pipeline.}
\label{fig:overview}
\end{figure*}

Weight-Space Programming (WSP) offers a more direct route. \citet{takehi2026weightspace} show that GPT-6 Astra, hereafter GPT-6, can read the weights of a neural network and edit them much like code. Asked to make \qwen better at math without gradients, it traced the small model's failures to a single direction in two MLP layers, removed that direction from 0.016\% of the weights, and raised MATH500 accuracy from 7.9\% to 48.1\%. 
Their experiments leave open whether a model can improve its performance by editing its own weights.

We study this possibility directly. In \swsp, \qwen edits itself. As Figure~\ref{fig:loop} shows, the model reads its own weights, proposes an edit, and continues with the edited weights, both as the solver that answers questions and as the editor that proposes the next edit. 
A fixed outer loop decides which edits to keep, and for this purpose we divide questions into three kinds. 
Practice questions come from the training split with reference answers, and are used to test whether an edit helps. 
Validation questions also come from the training split but are never shown to the editor, and serve only to select models after the search. 
The test sets are the standard GSM8K and MATH500 test questions, used only for the final evaluation. 
With this setup we ask four questions, and Figure~\ref{fig:overview} summarizes the test accuracy of the best model of each method.

\begin{itemize}
\item \textbf{Can a model improve itself through its weights?} Yes. Self-edits raise GSM8K accuracy about fivefold and MATH500 accuracy more than threefold, sometimes by changing a handful of numbers.
\item \textbf{Does the model know what it is doing?} Hardly. It edits whichever tensors are listed first, reuses numbers it has just read, and often gives hallucinated explanations. Many of the edits it keeps make the model worse.
\item \textbf{Can random edits do the same?} To some extent. Random search matches self-editing on GSM8K but falls short on MATH500, and the best single random edit is as good as the simpler of GPT-6's two published edits.
\item \textbf{Can \swsp keep improving the model?} In lucky cases. Continued self-editing lifts the best models further, to more than four times the original MATH500 accuracy, while typical runs make little further progress.
\end{itemize}

\noindent Self-improvement through the weights is therefore within reach of even a 0.8B model. Its current successes, however, come mostly from search and selection rather than from understanding, and closing this gap is the main challenge ahead.

\section{Self-Weight-Space Programming}
\label{sec:method}

\swsp is a loop in which a single model alternates between editing its weights and using them.

\paragraph{One model, two roles.}
Both roles are played by \qwen \citep{qwen2026qwen35}, a 24-layer model in which three of every four layers use linear attention \citep{yang2025gated} and the fourth uses standard attention. A linear-attention layer stores what it has read in a fixed-size state and forgets old information at a rate set by its weights. As the editor, the model proposes edits with thinking turned off. As the solver, it answers math questions with thinking turned on, using the sampling settings of \citet{takehi2026weightspace}. Both roles always use the current weights, so every kept edit changes both how the model solves problems and how it then proposes the next edit.

\paragraph{Reading its own weights.}
The editor sees a catalog of the 318 tensors it may edit, which covers all 24 layers and gives the name and shape of each tensor. After it picks one or two tensors, it receives their shape, statistics such as the mean and standard deviation, up to 64 of their values with positions, and the norms of their rows and columns. It also sees how the current model does on a batch of practice questions, including the reference answers and the end of the model's responses for a few examples. This matches the setting of \citet{takehi2026weightspace}, where GPT-6 also reads the actual weight values, can check its edits only on practice questions from the training split, and never sees the test sets. The difference is that GPT-6 could analyze the weights freely in a Python environment, whereas our editor sees only the information above, precomputed by the system. Appendix~\ref{app:interface} describes the full interface.

\paragraph{Writing its own weights.}
A model this small does not reliably follow output formats, so the editor writes its edits as JSON under a grammar built from the tensor shapes \citep{willard2023outlines,geng2023grammar}. Each proposal contains one or two edits. An edit selects a block of consecutive entries, rows or columns in a tensor and applies one of seven operations to it. It can set the block to zero, multiply it by a number, add a number, fill it with a number, redraw it at random from the tensor's mean and standard deviation, remove a direction, or apply a short NumPy expression. Removing a direction applies only to matrices. The editor names one column of the matrix as the direction, and every selected column subtracts its component along that direction, scaled by a coefficient. A coefficient of 1 removes this component entirely. GPT-6's most effective edit also removes a direction, but computes it from the model's activations. The grammar guarantees the format but not the content, and edits that repeat an earlier one, rescale a whole tensor uniformly or change no value are discarded at once. We call the remaining edits valid edits.

\paragraph{Keeping edits.}
Each run proceeds in rounds from a starting model. Every round has a fixed budget, the largest number of valid edits it may try, and the editor proposes edits until the budget is spent. A round also ends early after 24 proposals or after several valid edits in a row are rejected. Each valid edit is first applied to a copy of the weights and then checked on two disjoint practice batches. It is kept only if the edited model solves enough additional questions on the two batches combined without doing worse on either, in which case the copy becomes the new current weights. The budget, the batch size and the required gain depend on the interface, which we describe next.

\paragraph{Two interfaces.}
We run the loop with two interfaces, which differ mainly in how the tensor catalog is ordered. The shuffled interface lists all tensors in a random order, and the editor picks tensors directly. It uses 32 practice questions per batch, keeps any edit that gains a single answer in total, and runs three rounds with budgets of 4, 6 and 12 valid edits. The ordered interface lists the tensors in layer order, and the editor first picks a layer and then tensors within it. It uses 128 questions per batch, requires a total gain of at least three answers, and spends a budget of 8 valid edits per round. Appendix~\ref{app:protocol} gives the full settings.

\paragraph{Selection and testing.}
After every round we score the current model on validation questions, without ever feeding the result back to the editor. The round with the largest validation gain gives the best model of each run. A setting is one combination of dataset, interface and starting point and contains 12 or 24 runs. In each setting we take the best models of the three runs with the largest validation gain, name them \mbox{Top-1}, \mbox{Top-2} and \mbox{Top-3}, and call them the selected models. Only the selected models are evaluated, once each, on the full GSM8K \citep{cobbe2021gsm8k} and MATH500 \citep{hendrycks2021math,lightman2024verify} test sets. Being the best three of 12 or 24 runs, they show the best case. To see what an ordinary run achieves, we also designate several runs by a fixed rule before the search and test them whatever their outcome. Because they are never selected, these runs act as random spot checks, and we therefore refer to them as spot-check runs.

\paragraph{Continued runs.}
To study continued self-improvement, we also let the selected models of the shuffled interface keep running under the ordered interface, and call these continued runs. We write +$k$ for $k$ rounds of ordered self-editing accumulated on top of a selected shuffled model, and Section~\ref{sec:long} gives the details.

\section[Can a Model Improve Itself Through Its Weights?]{Can a Model Improve Itself\newline Through Its Weights?}
\label{sec:exist}

\begin{table}[t]
\centering
\small
\setlength{\tabcolsep}{4.2pt}
\begin{tabular}{@{}llrrrr@{}}
\toprule
 & Setting & Original & Top-1 & Top-2 & Top-3 \\
\midrule
\multirow{3}{*}{\rotatebox{90}{\scriptsize GSM8K}}
 & Shuffled & 12.5 & 31.9 & 28.7 & 16.5 \\
 & Ordered & 11.8 & \textbf{61.0} & 59.9 & 54.0 \\
 & Continued +8 & 11.8 & \textbf{62.2} & 53.3 & 48.1 \\
\midrule
\multirow{5}{*}{\rotatebox{90}{\scriptsize MATH500}}
 & Shuffled, batch 1 & 6.9 & 17.3 & 10.4 & 8.1\ns \\
 & Shuffled, batch 2 & 7.0 & 17.7 & 22.0 & 8.3\ns \\
 & Ordered & 7.0 & 24.2 & 21.9 & \textbf{24.7} \\
 & Continued +4 & 7.0 & 25.4 & 25.4 & 28.4 \\
 & Continued +12 & 7.0 & 31.8 & \textbf{32.2} & 28.6 \\
\bottomrule
\end{tabular}
\caption{Test accuracy in \% of all 24 selected models, ranked within each row by validation gain. Shuffled and ordered rows start from the original model. Continued rows start from selected models of the shuffled interface and keep running under the ordered interface, with +$k$ counting the accumulated ordered rounds, so the MATH500 +12 models come from eight more rounds on the selected +4 models. On MATH500 the shuffled interface ran in two batches of 24 runs. Original is the unedited model evaluated on the same hardware and sampling seeds. All gains over the original model are significant except those marked \dag.}
\label{tab:examples}
\end{table}

It can. Of the 24 selected models in Table~\ref{tab:examples}, 22 are significantly better than the original model, and the remaining two gain just over one point. Under the ordered interface, self-editing lifts GSM8K accuracy from 11.8\% to 61.0\% and MATH500 accuracy from 7.0\% to 24.7\%. Continued runs raise them further to 62.2\% and 32.2\%. The best self-edited GSM8K model comes within five points of GPT-6's best edit, which reaches 66.9\% in our evaluation, whereas on MATH500 GPT-6's best edit stays far ahead at 49.5\%.

\paragraph{A few numbers suffice.}
The successful edits are remarkably small. The 24.2\% MATH500 model differs from the original model in only 16 numbers, the \alog of the first linear-attention layer, which sets how quickly each head of that layer forgets the past. A GSM8K model that changes 32 numbers in the same layer reaches 59.9\%. One continued MATH500 model differs from its starting point in just two numbers yet gains another 6.8 points. No selected model changes more than 0.2\% of its weights, and Appendix~\ref{app:edits} lists every kept edit.

\section[Does the Model Know What It Is Doing?]{Does the Large Language Model Know What It Is Doing?}
\label{sec:anatomy}

The model clearly improves itself. How it does so is less flattering.

\begin{figure}[t]
\begin{tcolorbox}[colback=black!2,colframe=black!22,boxrule=0.4pt,arc=1.5pt,left=4pt,right=4pt,top=3pt,bottom=3pt]
\footnotesize
\renewcommand{\arraystretch}{1.12}
\begin{tabular}{@{}>{\bfseries}l@{\hspace{7pt}}>{\raggedright\arraybackslash}p{0.79\linewidth}@{}}
\color{inkgray}State & Two GSM8K practice batches of 128, with 12 and 20 correct \\
\color{navy}Call 1 & \texttt{\{"stop": false, "layer": 0\}} \\
\color{navy}Call 2 & \texttt{\{"tensors": ["p273", "p298"]\}}, the layer-0 input norm and \texttt{A\_log} \\
\color{inkgray}Read & Statistics of \texttt{A\_log}, including mean $-1.9795$ and standard deviation 1.4477 \\
\color{navy}Call 3 & \texttt{\{"tensor": "p273", "op": "scale", "count": 16, "value": -1.979492\}} \\
 & \texttt{\{"tensor": "p298", "op": "add", "count": 16, "value": -1.979492\}} \\
 & \texttt{"hypothesis":\newline "The current weights are scaled by -1.9795 ... an unintended scaling that should be removed"} \\
\color{inkgray}Result & Correct answers rise from 12 to 78 and from 20 to 87 on the two batches, so the edit is kept. The final model scores 61.0\% on the GSM8K test set. \\
\end{tabular}
\end{tcolorbox}
\vspace{-8pt}
\caption{The kept proposal behind the 61.0\% GSM8K model, abridged. The editor reuses the mean it has just read in both edits, while its explanation claims to undo a scaling by that same number. The catalog assigns each tensor an opaque identifier such as \code{p273} or \code{p298}, as described in Appendix~\ref{app:interface}.}
\label{fig:example}
\end{figure}

\paragraph{A lucky edit.}
Figure~\ref{fig:example} shows the key proposal behind the best GSM8K model started from the original model. The editor picks layer 0, reads the first two tensors of that layer, multiplies the first by the mean $-1.979$ of the second, and adds the same number to the second. Its explanation claims that these weights had been multiplied by $-1.9795$ by mistake and that this scaling should be removed. Undoing a multiplication by $-1.9795$, however, would require dividing by that number, yet the editor multiplies by it once more. 
The scaling it claims to remove is exactly the one it now applies, so the explanation contradicts the edit itself. 
The edit nevertheless raises the correct answers on the two practice batches from 12 and 20 to 78 and 87, and the resulting model scores 61.0\% on the test set.

\paragraph{Most proposals fail.}
This proposal is typical in everything but its luck. Of the 11,580 proposals the editor made, 69\% are invalid even though the grammar guarantees their format. Most of them repeat an edit already tried or rescale a whole tensor uniformly, which the prompt explicitly forbids. Only 12\% of the valid edits are kept. The editor never uses the two operations that can express structured changes, removing a direction and writing an expression, and almost always just multiplies or shifts a block of numbers.

\begin{figure}[t]
\centering
\includegraphics[width=\columnwidth]{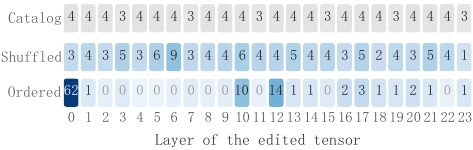}
\caption{Layer of every valid edit. Each cell gives the rounded percentage of valid edits that fall in that layer. The first row, in grey, is each layer's share of the catalog, the distribution an editor would follow if it treated all tensors alike. Under the shuffled interface, which lists the catalog in a random order, edits spread over the layers roughly in this proportion. Under the ordered interface, which lists the catalog by layer, 62\% of valid edits fall in layer 0, the first entry of the catalog.}
\label{fig:layers}
\end{figure}

\paragraph{Where it edits follows the catalog.}
With a shuffled catalog, the editor spreads its edits over all layers, and layer 0 receives 2.9\% of them, close to its 4.4\% share of the catalog. With a catalog ordered by layer, layer 0 receives 62\%, as Figure~\ref{fig:layers} shows. This preference appears already in the first proposal of each round, before the editor has seen the outcome of any proposal in that round. The first two tensors of layer 0 are exactly the two edited in Figure~\ref{fig:example}. All 15 selected models from the ordered interface edit \alog in layer 0, while none of the nine from the shuffled interface do. Changing how fast the first layer forgets is a reasonable target, but the editor seems to have arrived at it simply by reading the catalog from the top.

\paragraph{Its numbers copy the statistics it reads.}
The numbers in its edits are often copied as well. Between 9\% and 28\% of the values in valid edits equal a statistic the editor has just read, such as a mean, a standard deviation or a norm. The layer-0 \alog mean of $-1.979$ alone appears as a scale or a shift in four selected models.

\paragraph{Its explanations are often hallucinated.}
The explanations attached to kept edits are no more reliable. Many are empty, describe a block other than the one edited, or have nothing to do with weights at all. They are hallucinations in the usual sense for language models, fluent text that is unfaithful to the input or to the facts \citep{ji2023hallucination}. The most extreme one reads, in full, ``The cost of one chair is \$11.'' This sentence is copied word for word from the end of a solver response to a chair-pricing question in the practice feedback of the same call, and has nothing to do with the edit it accompanies.

\begin{figure}[t]
\centering
\includegraphics[width=0.96\columnwidth]{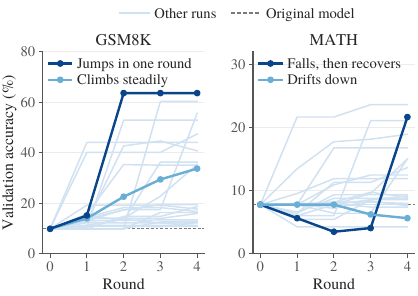}
\vspace{-8pt}
\caption{Validation accuracy of all runs started from the original model under the ordered interface, 24 per dataset. Two runs per dataset are highlighted to represent typical trajectories, thin light lines show the other runs, and the dashed line marks the original model.}
\label{fig:trajectories}
\end{figure}

\paragraph{Kept edits often hurt.}
Doing better on practice questions does not guarantee a better model. Estimating a model's ability from a small set of questions saves time \citep{huang2025minilongbench,ling2026survey}, but the estimate has a long-tailed error and can occasionally land far from the full evaluation \citep{ling2026bestrand}. Because the outer loop decides from practice scores alone, it inevitably keeps some edits that pass by luck. Under the shuffled interface, more than 40\% of the rounds that keep an edit lower validation accuracy. The larger batches of the ordered interface help, yet a third of such rounds on MATH still do. Figure~\ref{fig:trajectories} shows the resulting variety. Some runs jump in a single round, some climb steadily, some first fall below the original model and then recover, and some drift downward. On GSM8K, eleven of the 24 runs gain more than 23 points while the other thirteen gain less than 10. Spot-check runs, which are never selected, better represent an ordinary run. The ordered interface has six per dataset, and on the test sets they gain on average 24.7 points on GSM8K and 7.7 points on MATH500, far less than the selected models.

The edits are thus the model's own, made from its own weights without any test information. Yet where it edits follows the catalog it is shown, what it writes copies what it reads, and the reasons it gives are often hallucinated. The large gains come from a few lucky proposals, and it is the outer loop that keeps them.

\section{Can Random Edits Do the Same?}
\label{sec:random}

If lucky proposals drive the gains, random edits might do as well. We test this by replacing the editor with a random sampler that calls no model.

\begin{figure}[t]
\centering
\includegraphics[width=\columnwidth]{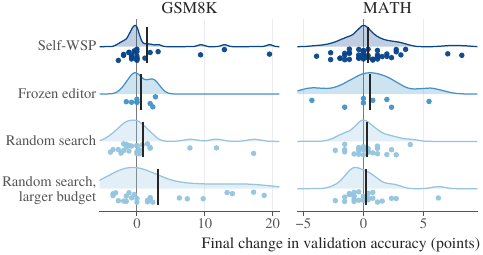}
\vspace{-15pt}
\caption{Final validation change of every run under the shuffled interface. Each dot is one run, the shaded curve above the dots shows their distribution, and the black bar marks the mean. The frozen editor always uses the original weights while only the solver is edited. Random search uses the same tensors, operations and selection rules, and its larger budget is about twice the self-editing budget on GSM8K and 1.7 times on MATH.}
\label{fig:random_search}
\end{figure}

\paragraph{Random search.}
For each proposal, the sampler draws one tensor uniformly from the catalog, or two with probability one fifth. It then picks a random layout, start and count, and a random operation with its value, choosing among setting to zero, scaling, adding or filling a small number, redrawing and removing a direction. Appendix~\ref{app:random} gives the exact distributions. The sampler runs through the same loop with the same selection rules as the shuffled interface. With the same budget, it spends exactly as many valid edits per round as the self-editing run with the same random seed. With the larger budget, it spends a fixed 4, 6 or 12 valid edits per round. On GSM8K, random search matches self-editing. Its best model scores 32.8\% against 31.9\% for self-editing, and Figure~\ref{fig:random_search} shows that the two sets of runs look much alike. We also test a frozen editor, which always uses the original weights while only the solver is edited. On average it behaves much like the other two, so letting the editor itself be edited brings no clear benefit. MATH500 is different. With the same budget, random search finds no improvement, while self-editing reaches 17.3\%. Only with about 70\% more budget does random search find a 13.6\% model, by zeroing 256 rows of one linear-attention layer.

\begin{figure}[t]
\centering
\includegraphics[width=\columnwidth]{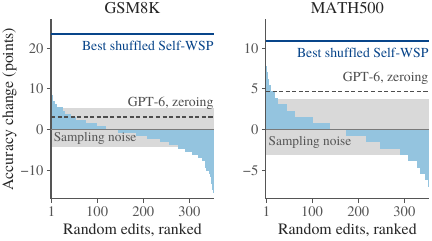}
\caption{The 352 single random edits applied to the original model, ranked from best to worst by the accuracy change they cause in the screening. The screening uses 128 held-out training questions with one sample each and an 8,192-token limit, so the vertical axis is meant for comparison within the figure and not against test accuracy. The grey band is the range of sampling noise when no edit is made. The solid blue line is the selected shuffled-interface model with the largest screening gain. Like the random edits, it starts from the original model, and its test accuracy is 31.9\% on GSM8K and 17.7\% on MATH500. The stronger models of the ordered interface were not part of this screening. The dashed line is GPT-6's simpler zeroing edit.}
\vspace{-10pt}
\label{fig:random_edits}
\end{figure}

\paragraph{Single random edits.}
Random search still selects and accumulates edits. To see what a single random edit does on its own, we remove both steps. For each dataset we draw 352 random edits, apply each alone to the original model, and evaluate it on 128 held-out training questions that are neither practice nor validation questions. We call this step the screening, and to save compute it samples each question once with an 8,192-token limit. Judging whether an edit really helps also requires knowing how much the score moves without any edit. The original model answers with some randomness, and merely answering again with another random seed moves its score on these 128 questions by a few points. We call this movement sampling noise and take its 5th to 95th percentile as the noise range, shown as the grey band in Figure~\ref{fig:random_edits}. An edit inside the band cannot be told apart from doing nothing.

As Figure~\ref{fig:random_edits} shows, most random edits fall inside the grey band or below it. On GSM8K, 78 edits fall below the band and only 24 rise above it. Large edits that touch tens of thousands to millions of weights do the most damage, and almost every kind of operation hurts on average. A few random edits nevertheless help. We rescore the best eight per dataset on 512 further questions from the practice pool and test the winner. It reaches 22.3\% on GSM8K by scaling 4,096 weights of a convolution by $-2$, and 14.0\% on MATH500 by scaling 64 rows of a query projection by $-2$. Appendix~\ref{app:random} gives the sampling details.

\paragraph{GPT-6's simpler edit is a random edit.}
One of GPT-6's two published edits sets 256 columns of the attention output matrix of layer 15 to zero. Such an edit lies within the range of the single random edits, although the sampler would draw it only about once in 130 million tries. In our evaluation it scores 22.0\% on GSM8K and 13.8\% on MATH500, as good as the best single random edits, and in the screening it merely ranks among the better random edits. GPT-6's stronger edit removes a direction computed from the model's activations, which lies outside what the sampler can produce.

\paragraph{Where self-editing beats chance.}
Self-editing does beat chance in places. In the same screening, the best self-edited model of the shuffled interface gains more than all 352 random edits on both datasets. On MATH500, the selected continued +4 models gain up to 17.2 points in the screening, more than twice the best random edit, and self-editing also beats random search with the same budget. Its proposals are thus better than random draws, though only modestly, and not because the model knows why they work.

\section[Can Self-WSP Keep Improving the Model over Time?]{Can Self-WSP Keep Improving the Model Round After Round?} 
\label{sec:long}

A model that truly improves itself should keep improving when given more time. We therefore let the selected models of the shuffled interface keep running under the ordered interface. On GSM8K, the three selected shuffled models, which we call A, B and C, each run eight more rounds, reaching +8. On MATH500, the six selected shuffled models first run four rounds each, reaching +4, and the three selected +4 models then run eight more rounds, reaching +12 in total. Each continued setting again yields three selected models. In the eight-round settings each starting model has four runs, and we designate the first in advance as a spot-check run, tested after its 4th and 8th rounds.

\begin{figure}[t]
\centering
\includegraphics[width=0.96\columnwidth]{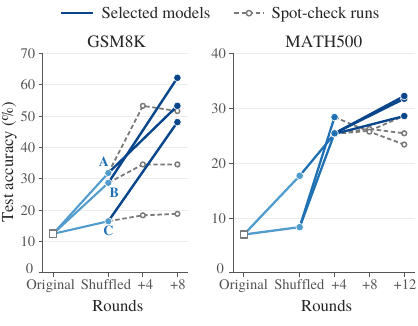}
\vspace{-10pt}
\caption{Continued self-editing. The horizontal axis shows the original model and the selected shuffled models, with +4, +8 and +12 marking the cumulative rounds of ordered self-editing from the shuffled models. Solid lines connect each selected model with its starting model. Each selected model comes from the round of its run with the best validation score and is drawn at the end of its setting. Dashed lines are spot-check runs. They start from the same models as the selected ones but are fixed before running and never selected, so they represent an ordinary run. Each is tested after its 4th and 8th rounds, which correspond to +4 and +8 on GSM8K and +8 and +12 on MATH500. Labels A, B and C mark the selected GSM8K models of the shuffled interface.}

\label{fig:chains}
\end{figure}

\paragraph{The best models keep climbing.}
The solid lines of Figure~\ref{fig:chains} show that the selected models do improve further. On GSM8K, model B rises from 28.7\% to 62.2\% at +8. On MATH500, a selected shuffled model at 8.3\% grows into models of 25.4\% to 28.4\% at +4, and one of these reaches 32.2\% at +12 while differing from its starting point in just two numbers. Most of these gains arrive soon after the switch to the ordered interface, when runs find the layer-0 edit of Section~\ref{sec:anatomy}.

\paragraph{Typical runs stall.}
The selected models are the best three of 12 runs and naturally favor the lucky ones. The spot-check runs, the dashed lines in Figure~\ref{fig:chains}, show what an ordinary run achieves, and they tell a different story. On GSM8K the spot-check runs gain little after +4, and the best of them slips from 53.3\% at +4 to 51.6\% at +8. On MATH500 all three start from selected +4 models, and by +12 one gains three points, one ends where it started, and one falls steadily from 28.4\% to 23.4\%. Across all 12 eight-round runs on MATH, kept edits leave validation accuracy unchanged on average.

More rounds give the outer loop more chances to find a good model, but although its weights keep changing, the model itself does not become any better as an editor.

\section{Discussion}
\label{sec:discussion}

\paragraph{The outer loop does much of the work.}
The model proposes every edit and is changed by every kept one, but the outer loop does the rest. Using practice questions with reference answers, it decides which edits to keep, when to stop and which run to test. \swsp is therefore best seen as a search in which the model supplies the proposals, and these proposals beat random ones on MATH500 but merely match them on GSM8K.

\paragraph{Why small edits go a long way.}
As its model card warns, \qwen often gets stuck repeating itself while thinking, and the original model fails to finish within its 32k-token budget on about 90\% of questions. Almost every answer that an edit makes correct is one the original model never finished, and across all evaluated models accuracy falls almost linearly with the share of unfinished answers. Any edit that breaks the repetition can therefore raise accuracy sharply. This fits reports that a few parameters can have outsized effects \citep{yu2024superweight}, that correcting a small fraction of thinking tokens during reasoning clearly raises accuracy \citep{ling2026ncots}, and that Qwen models improve at math even when trained with spurious rewards \citep{shao2025spurious}. It may also explain why GPT-6 could not improve Gemma 3 270M, a model that rarely repeats itself \citep{takehi2026weightspace}.

\paragraph{Toward genuine self-improvement.}
A model that improves itself because it understands its weights must act on what it reads. Natural next steps include tools for analyzing weights, which GPT-6 had and our editor lacked, training a model to propose edits \citep{zweiger2025seal}, and solvers whose failures run deeper than repetition. Progress along this path is best measured against random search with the same interface and budget, and reported together with spot-check runs.

\section{Related Work}
\label{sec:related}

\paragraph{Large language models.}
Scaling models and data has given language models broad abilities \citep{kaplan2020scaling,brown2020gpt3,touvron2023llama,openai2023gpt4,yang2025qwen3}.
Chain-of-thought prompting elicits intermediate reasoning steps \citep{wei2022cot}.
Reinforcement learning trains models to reason at length \citep{deepseek2025r1}, and test-time scaling trades more inference compute for higher accuracy \citep{snell2025scaling,wan2026tts}, for example by searching for better reasoning paths \citep{ling2026ncots}.
Mathematical reasoning is a common testbed for testing and improving these abilities \citep{cobbe2021gsm8k,hendrycks2021math,qin2024idam}.
These advances rely on training, prompting or extra inference compute, whereas \swsp lets a model directly rewrite its own weights without gradients.

\paragraph{Self-improvement.}
Models can refine their outputs with their own feedback \citep{madaan2023selfrefine} or generate more creative data and train on the best of it \citep{zhong2024clot}, although they often fail to correct their own reasoning \citep{huang2024cannot}.
Other work lets a model rewrite its own prompts, agent designs or running logic and keeps the better versions \citep{fernando2024promptbreeder,hu2025adas,yin2025godelagent}. All of these methods act on data, prompts or code, whereas \swsp acts on the weights themselves. The idea of a neural network that modifies its own weights is old \citep{schmidhuber1993selfreferential}, and modern self-referential weight matrices can rewrite themselves at run time \citep{irie2022modern}. These networks, however, are trained for this purpose, while \swsp lets a general-purpose language model read and write its own weights from instructions in natural language.

\paragraph{Editing and writing weights.}
Model editing changes facts or behaviors through targeted weight updates \citep{meng2022rome,meng2023memit}, trained editor networks \citep{mitchell2022mend}, arithmetic on weight differences \citep{ilharco2023taskarithmetic} or the removal of single directions \citep{arditi2024refusal}, usually guided by interpretability \citep{elhage2021mathematical}. Pruning research uses weight statistics such as norms to judge which parameters can be removed, finds that these statistics often fail to tell parameters apart \citep{huang2021rethinking}, and turns to search to blend several criteria \citep{he2021blending}. Hypernetworks use one network to generate the weights of another \citep{ha2017hypernetworks}, and language models have been used to explain the neurons of other models \citep{bills2023language}. WSP instead asks a general-purpose language model to read and write weights toward a goal stated in natural language \citep{takehi2026weightspace}.
\citet{fan2026language} improved Qwen3.5-27B's math accuracy using a frozen copy as editor.
Other WSP studies include \citet{kundu2026same,hu2026exploring,fischetti2026just,salloum2026tree}.

\paragraph{Propose and select.}
Our loop resembles methods in which a language model proposes candidates and an evaluator keeps the best. Such methods have been applied to prompts, pipelines, programs and algorithms \citep{yang2024opro,khattab2024dspy,romeraparedes2024funsearch,novikov2025alphaevolve}, and a language model can even act directly as an evolution strategy over numerical parameters \citep{lange2024evollm}. Evolutionary search has also been used to find ways of merging the weights of several models \citep{akiba2025evolutionary}. In \swsp, the language model that proposes the candidates is itself the object being optimized.

\paragraph{Gradient-free and random baselines.}
Evolution strategies and zeroth-order methods optimize networks with forward passes alone \citep{salimans2017es,malladi2023mezo}, now even for language models with billions of parameters \citep{qiu2025es}. Random search is a strong baseline in hyperparameter optimization and reinforcement learning \citep{bergstra2012random,mania2018random}, and in benchmark compression a randomly drawn subset of questions often comes close to the full benchmark \citep{ling2026bestrand}. Repetition and overthinking are known failures of text generation and reasoning models \citep{holtzman2020degeneration,xu2022breakloop,chen2024overthinking}. Our results connect these threads. When models fail mainly through repetition, random perturbations can also be useful.

\section{Conclusion}

We asked whether a language model can improve itself by editing its own weights. Our results show that a 0.8B model can. Its own edits raise GSM8K accuracy fivefold and MATH500 accuracy more than fourfold, sometimes by changing a handful of numbers, and in lucky runs continued self-editing keeps paying off. 
Despite the improvements, the model shows little understanding of its edits, and random edits sometimes achieve comparable gains. \swsp is a promising path toward RSI, and turning it into a reliable ability is the work ahead.

\section*{Limitations}

We study a single small model on two math benchmarks, and its main weakness, repetitive thinking, is unusually easy to repair. The outer loop uses the reference answers of practice questions to decide which edits to keep, so the improvement is supervised. The ordered interface was designed after we had seen the results of the shuffled one. We have not yet run random search under it, nor tested the source of the layer-0 preference directly by reversing the catalog. Finally, the comparison with GPT-6 is only indicative, since GPT-6 used different tools, feedback and budgets.

\section*{Ethics Statement}
Self-editing of weights may introduce behaviors that a math benchmark does not measure. All edits in this study are logged and reproducible and concern a 0.8B model evaluated on public benchmarks. We use \qwen under the Apache-2.0 license, GSM8K and MATH under the MIT license, and the GPT-6 edits under CC BY 4.0.

\bibliography{custom}

@misc{takehi2026weightspace,
  title        = {Sparks of Weight-Space Programming: {LLMs} can natively read \& write neural network weights},
  author       = {Takehi, Rikiya and Teoh, Jayden and Zhang, Alex and Khattab, Omar},
  year         = {2026},
  month        = nov,
  howpublished = {\url{https://rikiyat.github.io/blog/weight-space-programming/}},
  note         = {Blog post}
}

@misc{qwen2026qwen35,
  title        = {{Qwen3.5-0.8B} Model Card},
  author       = {{Qwen Team}},
  year         = {2026},
  howpublished = {\url{https://huggingface.co/Qwen/Qwen3.5-0.8B}}
}

@article{yang2025qwen3,
  title   = {{Qwen3} Technical Report},
  author  = {Yang, An and Li, Anfeng and Yang, Baosong and Zhang, Beichen and Hui, Binyuan and Zheng, Bo and Yu, Bowen and Gao, Chang and Huang, Chengen and Lv, Chenxu and others},
  journal = {arXiv preprint arXiv:2505.09388},
  year    = {2025}
}

@inproceedings{yang2025gated,
  title     = {Gated Delta Networks: Improving {Mamba2} with Delta Rule},
  author    = {Yang, Songlin and Kautz, Jan and Hatamizadeh, Ali},
  booktitle = {The Thirteenth International Conference on Learning Representations},
  year      = {2025}
}

@article{cobbe2021gsm8k,
  title   = {Training Verifiers to Solve Math Word Problems},
  author  = {Cobbe, Karl and Kosaraju, Vineet and Bavarian, Mohammad and Chen, Mark and Jun, Heewoo and Kaiser, Lukasz and Plappert, Matthias and Tworek, Jerry and Hilton, Jacob and Nakano, Reiichiro and Hesse, Christopher and Schulman, John},
  journal = {arXiv preprint arXiv:2110.14168},
  year    = {2021}
}

@inproceedings{hendrycks2021math,
  title     = {Measuring Mathematical Problem Solving With the {MATH} Dataset},
  author    = {Hendrycks, Dan and Burns, Collin and Kadavath, Saurav and Arora, Akul and Basart, Steven and Tang, Eric and Song, Dawn and Steinhardt, Jacob},
  booktitle = {Proceedings of the Neural Information Processing Systems Track on Datasets and Benchmarks},
  year      = {2021}
}

@inproceedings{lightman2024verify,
  title     = {Let's Verify Step by Step},
  author    = {Lightman, Hunter and Kosaraju, Vineet and Burda, Yura and Edwards, Harri and Baker, Bowen and Lee, Teddy and Leike, Jan and Schulman, John and Sutskever, Ilya and Cobbe, Karl},
  booktitle = {The Twelfth International Conference on Learning Representations},
  year      = {2024}
}

@incollection{good1966speculations,
title = {Speculations Concerning the First Ultraintelligent Machine},
editor = {Franz L. Alt and Morris Rubinoff},
booktitle = {Advances in Computers},
publisher = {Elsevier},
volume = {6},
pages = {31-88},
year = {1966},
issn = {0065-2458},
doi = {https://doi.org/10.1016/S0065-2458(08)60418-0},
url = {https://www.sciencedirect.com/science/article/pii/S0065245808604180},
author = {Irving John Good}
}

@incollection{schmidhuber2007godel,
  title     = {{G}{\"o}del Machines: Fully Self-Referential Optimal Universal Self-Improvers},
  author    = {Schmidhuber, J{\"u}rgen},
  booktitle = {Artificial General Intelligence},
  pages     = {199--226},
  publisher = {Springer},
  year      = {2007}
}

@inproceedings{zelikman2022star,
 author = {Zelikman, Eric and Wu, Yuhuai and Mu, Jesse and Goodman, Noah},
 booktitle = {Advances in Neural Information Processing Systems},
 doi = {10.52202/068431-1126},
 editor = {S. Koyejo and S. Mohamed and A. Agarwal and D. Belgrave and K. Cho and A. Oh},
 pages = {15476--15488},
 publisher = {Curran Associates, Inc.},
 title = {STaR: Bootstrapping Reasoning With Reasoning},
 url = {https://proceedings.neurips.cc/paper_files/paper/2022/file/639a9a172c044fbb64175b5fad42e9a5-Paper-Conference.pdf},
 volume = {35},
 year = {2022}
}

@inproceedings{huang2023selfimprove,
  title     = {Large Language Models Can Self-Improve},
  author    = {Huang, Jiaxin and Gu, Shixiang Shane and Hou, Le and Wu, Yuexin and Wang, Xuezhi and Yu, Hongkun and Han, Jiawei},
  booktitle = {Proceedings of the 2023 Conference on Empirical Methods in Natural Language Processing},
  pages     = {1051--1068},
  year      = {2023}
}

@inproceedings{yuan2024selfrewarding,
  title = 	 {Self-Rewarding Language Models},
  author =       {Yuan, Weizhe and Pang, Richard Yuanzhe and Cho, Kyunghyun and Li, Xian and Sukhbaatar, Sainbayar and Xu, Jing and Weston, Jason E},
  booktitle = 	 {Proceedings of the 41st International Conference on Machine Learning},
  pages = 	 {57905--57923},
  year = 	 {2024},
  editor = 	 {Salakhutdinov, Ruslan and Kolter, Zico and Heller, Katherine and Weller, Adrian and Oliver, Nuria and Scarlett, Jonathan and Berkenkamp, Felix},
  volume = 	 {235},
  series = 	 {Proceedings of Machine Learning Research},
  month = 	 {21--27 Jul},
  publisher =    {PMLR},
  url = 	 {https://proceedings.mlr.press/v235/yuan24d.html}
}

@inproceedings{zelikman2024stop,
  title     = {Self-Taught Optimizer ({STOP}): Recursively Self-Improving Code Generation},
  author    = {Zelikman, Eric and Lorch, Eliana and Mackey, Lester and Kalai, Adam Tauman},
  booktitle = {First Conference on Language Modeling},
  year      = {2024}
}

@inproceedings{zweiger2025seal,
 author = {Zweiger, Adam and Pari, Jyo and Guo, Han and Kim, Yoon and Agrawal, Pulkit},
 booktitle = {Advances in Neural Information Processing Systems},
 doi = {10.52202/085713-2483},
 editor = {D. Belgrave and C. Zhang and H. Lin and R. Pascanu and P. Koniusz and M. Ghassemi and N. Chen},
 pages = {74084--74115},
 publisher = {Curran Associates, Inc.},
 title = {Self-Adapting Language Models},
 url = {https://proceedings.neurips.cc/paper_files/paper/2025/file/6b41e04c41726e2a60e456d0a2b961ab-Paper-Conference.pdf},
 volume = {38, Main Conference},
 year = {2025}
}

@inproceedings{zhang2025dgm,
  title     = {Darwin {G}{\"o}del Machine: Open-Ended Evolution of Self-Improving Agents},
  author    = {Zhang, Jenny and Hu, Shengran and Lu, Cong and Lange, Robert and Clune, Jeff},
  booktitle = {The Fourteenth International Conference on Learning Representations},
  year      = {2026}
}

@inproceedings{huang2024cannot,
  title     = {Large Language Models Cannot Self-Correct Reasoning Yet},
  author    = {Huang, Jie and Chen, Xinyun and Mishra, Swaroop and Zheng, Huaixiu Steven and Yu, Adams Wei and Song, Xinying and Zhou, Denny},
  booktitle = {The Twelfth International Conference on Learning Representations},
  year      = {2024}
}

@inproceedings{yang2024opro,
  title     = {Large Language Models as Optimizers},
  author    = {Yang, Chengrun and Wang, Xuezhi and Lu, Yifeng and Liu, Hanxiao and Le, Quoc V. and Zhou, Denny and Chen, Xinyun},
  booktitle = {The Twelfth International Conference on Learning Representations},
  year      = {2024}
}

@inproceedings{khattab2024dspy,
title={{DSP}y: Compiling Declarative Language Model Calls into State-of-the-Art Pipelines},
author={Omar Khattab and Arnav Singhvi and Paridhi Maheshwari and Zhiyuan Zhang and Keshav Santhanam and Sri Vardhamanan A and Saiful Haq and Ashutosh Sharma and Thomas T. Joshi and Hanna Moazam and Heather Miller and Matei Zaharia and Christopher Potts},
booktitle={The Twelfth International Conference on Learning Representations},
year={2024},
url={https://openreview.net/forum?id=sY5N0zY5Od}
}

@article{romeraparedes2024funsearch,
  title   = {Mathematical Discoveries from Program Search with Large Language Models},
  author  = {Romera-Paredes, Bernardino and Barekatain, Mohammadamin and Novikov, Alexander and Balog, Matej and Kumar, M. Pawan and Dupont, Emilien and Ruiz, Francisco J. R. and Ellenberg, Jordan S. and Wang, Pengming and Fawzi, Omar and Kohli, Pushmeet and Fawzi, Alhussein},
  journal = {Nature},
  volume  = {625},
  pages   = {468--475},
  year    = {2024}
}

@inproceedings{meng2022rome,
 author = {Meng, Kevin and Bau, David and Andonian, Alex and Belinkov, Yonatan},
 booktitle = {Advances in Neural Information Processing Systems},
 doi = {10.52202/068431-1262},
 editor = {S. Koyejo and S. Mohamed and A. Agarwal and D. Belgrave and K. Cho and A. Oh},
 pages = {17359--17372},
 publisher = {Curran Associates, Inc.},
 title = {Locating and Editing Factual Associations in GPT},
 url = {https://proceedings.neurips.cc/paper_files/paper/2022/file/6f1d43d5a82a37e89b0665b33bf3a182-Paper-Conference.pdf},
 volume = {35},
 year = {2022}
}

@inproceedings{meng2023memit,
title={Mass-Editing Memory in a Transformer},
author={Kevin Meng and Arnab Sen Sharma and Alex J Andonian and Yonatan Belinkov and David Bau},
booktitle={The Eleventh International Conference on Learning Representations },
year={2023},
url={https://openreview.net/forum?id=MkbcAHIYgyS}
}

@inproceedings{ilharco2023taskarithmetic,
  title     = {Editing Models with Task Arithmetic},
  author    = {Ilharco, Gabriel and Ribeiro, Marco Tulio and Wortsman, Mitchell and Gururangan, Suchin and Schmidt, Ludwig and Hajishirzi, Hannaneh and Farhadi, Ali},
  booktitle = {The Eleventh International Conference on Learning Representations},
  year      = {2023}
}

@inproceedings{arditi2024refusal,
  title     = {Refusal in Language Models Is Mediated by a Single Direction},
  author    = {Arditi, Andy and Obeso, Oscar and Syed, Aaquib and Paleka, Daniel and Panickssery, Nina and Gurnee, Wes and Nanda, Neel},
  booktitle = {Advances in Neural Information Processing Systems},
  volume    = {37},
  year      = {2024}
}

@article{yu2024superweight,
  title   = {The Super Weight in Large Language Models},
  author  = {Yu, Mengxia and Wang, De and Shan, Qi and Reed, Colorado J. and Wan, Alvin},
  journal = {arXiv preprint arXiv:2411.07191},
  year    = {2024}
}

@article{salimans2017es,
  title   = {Evolution Strategies as a Scalable Alternative to Reinforcement Learning},
  author  = {Salimans, Tim and Ho, Jonathan and Chen, Xi and Sidor, Szymon and Sutskever, Ilya},
  journal = {arXiv preprint arXiv:1703.03864},
  year    = {2017}
}

@inproceedings{malladi2023mezo,
  title     = {Fine-Tuning Language Models with Just Forward Passes},
  author    = {Malladi, Sadhika and Gao, Tianyu and Nichani, Eshaan and Damian, Alex and Lee, Jason D. and Chen, Danqi and Arora, Sanjeev},
  booktitle = {Advances in Neural Information Processing Systems},
  volume    = {36},
  year      = {2023}
}

@inproceedings{qiu2025es,
  title = 	 {Evolution Strategies at Scale: {LLM} Fine-Tuning Beyond Reinforcement Learning},
  author =       {Qiu, Xin and Gan, Yulu and Hayes, Conor F. and Liang, Qiyao and Xu, Yinggan and Dailey, Roberto and Meyerson, Elliot and Hodjat, Babak and Miikkulainen, Risto},
  booktitle = 	 {Proceedings of the 43rd International Conference on Machine Learning},
  pages = 	 {101987--102015},
  year = 	 {2026},
  editor = 	 {Zhang, Tong and Dudik, Miroslav and Jaggi, Martin and Agarwal, Alekh and Li, Sharon and Schuurmans, Dale and Zhu, Jerry and Berkenkamp, Felix and Dong, Hanze and Bietti, Alberto},
  volume = 	 {306},
  series = 	 {Proceedings of Machine Learning Research},
  month = 	 {06--11 Jul},
  publisher =    {PMLR},
  url = 	 {https://proceedings.mlr.press/v306/qiu26f.html},
}

@inproceedings{mania2018random,
  title     = {Simple Random Search of Static Linear Policies is Competitive for Reinforcement Learning},
  author    = {Mania, Horia and Guy, Aurelia and Recht, Benjamin},
  booktitle = {Advances in Neural Information Processing Systems},
  volume    = {31},
  year      = {2018}
}

@article{bergstra2012random,
  title   = {Random Search for Hyper-Parameter Optimization},
  author  = {Bergstra, James and Bengio, Yoshua},
  journal = {Journal of Machine Learning Research},
  volume  = {13},
  pages   = {281--305},
  year    = {2012}
}

@inproceedings{shao2025spurious,
  title     = {Spurious Rewards: Rethinking Training Signals in {RLVR}},
  author    = {Shao, Rulin and Li, Shuyue Stella and Xin, Rui and Geng, Scott and Wang, Yiping and Oh, Sewoong and Du, Simon Shaolei and Lambert, Nathan and Min, Sewon and Krishna, Ranjay and Tsvetkov, Yulia and Hajishirzi, Hannaneh and Koh, Pang Wei and Zettlemoyer, Luke},
  booktitle = {Proceedings of the 43rd International Conference on Machine Learning},
  series    = {Proceedings of Machine Learning Research},
  volume    = {306},
  pages     = {110167--110207},
  year      = {2026}
}

@inproceedings{holtzman2020degeneration,
  title     = {The Curious Case of Neural Text Degeneration},
  author    = {Holtzman, Ari and Buys, Jan and Du, Li and Forbes, Maxwell and Choi, Yejin},
  booktitle = {International Conference on Learning Representations},
  year      = {2020}
}

@inproceedings{chen2024overthinking,
  title     = {Do {NOT} Think That Much for 2+3=? On the Overthinking of Long Reasoning Models},
  author    = {Chen, Xingyu and Xu, Jiahao and Liang, Tian and He, Zhiwei and Pang, Jianhui and Yu, Dian and Song, Linfeng and Liu, Qiuzhi and Zhou, Mengfei and Zhang, Zhuosheng and Wang, Rui and Tu, Zhaopeng and Mi, Haitao and Yu, Dong},
  booktitle = {Proceedings of the 42nd International Conference on Machine Learning},
  series    = {Proceedings of Machine Learning Research},
  volume    = {267},
  pages     = {9487--9499},
  year      = {2025}
}

@article{deepseek2025r1,
  title   = {{DeepSeek-R1} Incentivizes Reasoning in {LLMs} through Reinforcement Learning},
  author  = {Guo, Daya and others},
  journal = {Nature},
  volume  = {645},
  number  = {8081},
  pages   = {633--638},
  year    = {2025}
}

@article{willard2023outlines,
  title   = {Efficient Guided Generation for Large Language Models},
  author  = {Willard, Brandon T. and Louf, R{\'e}mi},
  journal = {arXiv preprint arXiv:2307.09702},
  year    = {2023}
}

@inproceedings{geng2023grammar,
  title     = {Grammar-Constrained Decoding for Structured {NLP} Tasks without Finetuning},
  author    = {Geng, Saibo and Josifoski, Martin and Peyrard, Maxime and West, Robert},
  booktitle = {Proceedings of the 2023 Conference on Empirical Methods in Natural Language Processing},
  pages     = {10932--10952},
  year      = {2023}
}

@article{elhage2021mathematical,
  title   = {A Mathematical Framework for Transformer Circuits},
  author  = {Elhage, Nelson and Nanda, Neel and Olsson, Catherine and Henighan, Tom and Joseph, Nicholas and Mann, Ben and Askell, Amanda and Bai, Yuntao and Chen, Anna and Conerly, Tom and DasSarma, Nova and Drain, Dawn and Ganguli, Deep and Hatfield-Dodds, Zac and Hernandez, Danny and Jones, Andy and Kernion, Jackson and Lovitt, Liane and Ndousse, Kamal and Amodei, Dario and Brown, Tom and Clark, Jack and Kaplan, Jared and McCandlish, Sam and Olah, Chris},
  journal = {Transformer Circuits Thread},
  year    = {2021}
}

@inproceedings{ha2017hypernetworks,
  title     = {{HyperNetworks}},
  author    = {Ha, David and Dai, Andrew M. and Le, Quoc V.},
  booktitle = {International Conference on Learning Representations},
  year      = {2017}
}

@inproceedings{madaan2023selfrefine,
  title     = {Self-Refine: Iterative Refinement with Self-Feedback},
  author    = {Madaan, Aman and Tandon, Niket and Gupta, Prakhar and Hallinan, Skyler and Gao, Luyu and Wiegreffe, Sarah and Alon, Uri and Dziri, Nouha and Prabhumoye, Shrimai and Yang, Yiming and Gupta, Shashank and Majumder, Bodhisattwa Prasad and Hermann, Katherine and Welleck, Sean and Yazdanbakhsh, Amir and Clark, Peter},
  booktitle = {Advances in Neural Information Processing Systems},
  volume    = {36},
  year      = {2023}
}

@inproceedings{schmidhuber1993selfreferential,
  title     = {A Self-Referential Weight Matrix},
  author    = {Schmidhuber, J{\"u}rgen},
  booktitle = {Proceedings of the International Conference on Artificial Neural Networks},
  pages     = {446--450},
  publisher = {Springer},
  year      = {1993}
}

@inproceedings{irie2022modern,
  title     = {A Modern Self-Referential Weight Matrix That Learns to Modify Itself},
  author    = {Irie, Kazuki and Schlag, Imanol and Csord{\'a}s, R{\'o}bert and Schmidhuber, J{\"u}rgen},
  booktitle = {Proceedings of the 39th International Conference on Machine Learning},
  series    = {Proceedings of Machine Learning Research},
  volume    = {162},
  pages     = {9660--9677},
  year      = {2022}
}

@inproceedings{mitchell2022mend,
  title     = {Fast Model Editing at Scale},
  author    = {Mitchell, Eric and Lin, Charles and Bosselut, Antoine and Finn, Chelsea and Manning, Christopher D.},
  booktitle = {International Conference on Learning Representations},
  year      = {2022}
}

@inproceedings{fernando2024promptbreeder,
  title     = {Promptbreeder: Self-Referential Self-Improvement via Prompt Evolution},
  author    = {Fernando, Chrisantha and Banarse, Dylan and Michalewski, Henryk and Osindero, Simon and Rockt{\"a}schel, Tim},
  booktitle = {Proceedings of the 41st International Conference on Machine Learning},
  series    = {Proceedings of Machine Learning Research},
  volume    = {235},
  pages     = {13481--13544},
  year      = {2024}
}

@article{novikov2025alphaevolve,
  title   = {{AlphaEvolve}: A Coding Agent for Scientific and Algorithmic Discovery},
  author  = {Novikov, Alexander and V{\~u}, Ng{\^a}n and Eisenberger, Marvin and Dupont, Emilien and Huang, Po-Sen and Wagner, Adam Zsolt and Shirobokov, Sergey and Kozlovskii, Borislav and Ruiz, Francisco J. R. and Mehrabian, Abbas and Kumar, M. Pawan and See, Abigail and Chaudhuri, Swarat and Holland, George and Davies, Alex and Nowozin, Sebastian and Kohli, Pushmeet and Balog, Matej},
  journal = {arXiv preprint arXiv:2506.13131},
  year    = {2025}
}

@inproceedings{yin2025godelagent,
  title     = {G{\"o}del Agent: A Self-Referential Agent Framework for Recursively Self-Improvement},
  author    = {Yin, Xunjian and Wang, Xinyi and Pan, Liangming and Lin, Li and Wan, Xiaojun and Wang, William Yang},
  booktitle = {Proceedings of the 63rd Annual Meeting of the Association for Computational Linguistics (Volume 1: Long Papers)},
  pages     = {27890--27913},
  year      = {2025}
}

@inproceedings{hu2025adas,
  title     = {Automated Design of Agentic Systems},
  author    = {Hu, Shengran and Lu, Cong and Clune, Jeff},
  booktitle = {The Thirteenth International Conference on Learning Representations},
  year      = {2025}
}

@article{akiba2025evolutionary,
  title   = {Evolutionary Optimization of Model Merging Recipes},
  author  = {Akiba, Takuya and Shing, Makoto and Tang, Yujin and Sun, Qi and Ha, David},
  journal = {Nature Machine Intelligence},
  volume  = {7},
  year    = {2025},
  doi     = {10.1038/s42256-024-00975-8}
}

@inproceedings{lange2024evollm,
  title     = {Large Language Models As Evolution Strategies},
  author    = {Lange, Robert Tjarko and Tian, Yingtao and Tang, Yujin},
  booktitle = {Proceedings of the Genetic and Evolutionary Computation Conference Companion},
  pages     = {579--582},
  year      = {2024}
}

@inproceedings{xu2022breakloop,
  title     = {Learning to Break the Loop: Analyzing and Mitigating Repetitions for Neural Text Generation},
  author    = {Xu, Jin and Liu, Xiaojiang and Yan, Jianhao and Cai, Deng and Li, Huayang and Li, Jian},
  booktitle = {Advances in Neural Information Processing Systems},
  volume    = {35},
  pages     = {3082--3095},
  year      = {2022}
}

@misc{bills2023language,
  title        = {Language Models Can Explain Neurons in Language Models},
  author       = {Bills, Steven and Cammarata, Nick and Mossing, Dan and Tillman, Henk and Gao, Leo and Goh, Gabriel and Sutskever, Ilya and Leike, Jan and Wu, Jeff and Saunders, William},
  year         = {2023},
  howpublished = {\url{https://openaipublic.blob.core.windows.net/neuron-explainer/paper/index.html}}
}

@article{ji2023hallucination,
  title   = {Survey of Hallucination in Natural Language Generation},
  author  = {Ji, Ziwei and Lee, Nayeon and Frieske, Rita and Yu, Tiezheng and Su, Dan and Xu, Yan and Ishii, Etsuko and Bang, Ye Jin and Madotto, Andrea and Fung, Pascale},
  journal = {ACM Computing Surveys},
  volume  = {55},
  number  = {12},
  pages   = {248:1--248:38},
  year    = {2023}
}

@article{kaplan2020scaling,
  title   = {Scaling Laws for Neural Language Models},
  author  = {Kaplan, Jared and McCandlish, Sam and Henighan, Tom and Brown, Tom B. and Chess, Benjamin and Child, Rewon and Gray, Scott and Radford, Alec and Wu, Jeffrey and Amodei, Dario},
  journal = {arXiv preprint arXiv:2001.08361},
  year    = {2020}
}

@inproceedings{brown2020gpt3,
  title     = {Language Models are Few-Shot Learners},
  author    = {Brown, Tom B. and Mann, Benjamin and Ryder, Nick and Subbiah, Melanie and Kaplan, Jared and Dhariwal, Prafulla and Neelakantan, Arvind and Shyam, Pranav and Sastry, Girish and Askell, Amanda and others},
  booktitle = {Advances in Neural Information Processing Systems},
  volume    = {33},
  pages     = {1877--1901},
  year      = {2020}
}

@article{touvron2023llama,
  title   = {{LLaMA}: Open and Efficient Foundation Language Models},
  author  = {Touvron, Hugo and Lavril, Thibaut and Izacard, Gautier and Martinet, Xavier and Lachaux, Marie-Anne and Lacroix, Timoth{\'e}e and Rozi{\`e}re, Baptiste and Goyal, Naman and Hambro, Eric and Azhar, Faisal and Rodriguez, Aurelien and Joulin, Armand and Grave, Edouard and Lample, Guillaume},
  journal = {arXiv preprint arXiv:2302.13971},
  year    = {2023}
}

@article{openai2023gpt4,
  title   = {{GPT-4} Technical Report},
  author  = {{OpenAI}},
  journal = {arXiv preprint arXiv:2303.08774},
  year    = {2023}
}

@inproceedings{wei2022cot,
  title     = {Chain-of-Thought Prompting Elicits Reasoning in Large Language Models},
  author    = {Wei, Jason and Wang, Xuezhi and Schuurmans, Dale and Bosma, Maarten and Ichter, Brian and Xia, Fei and Chi, Ed and Le, Quoc V. and Zhou, Denny},
  booktitle = {Advances in Neural Information Processing Systems},
  volume    = {35},
  pages     = {24824--24837},
  year      = {2022}
}

@inproceedings{snell2025scaling,
  title     = {Scaling {LLM} Test-Time Compute Optimally Can be More Effective than Scaling Parameters for Reasoning},
  author    = {Snell, Charlie and Lee, Jaehoon and Xu, Kelvin and Kumar, Aviral},
  booktitle = {The Thirteenth International Conference on Learning Representations},
  year      = {2025}
}

@inproceedings{wan2026tts,
  title     = {Test-Time Scaling in Multimodal Foundation Models: A Comprehensive Survey of Generation and Reasoning},
  author    = {Wan, Cong and He, Ying and Huang, Zhongzhan and Wu, Hefeng},
  booktitle = {Findings of the Association for Computational Linguistics: ACL 2026},
  pages     = {7751--7767},
  year      = {2026}
}

@inproceedings{ling2026ncots,
  title     = {Neural Chain-of-Thought Search: Searching the Optimal Reasoning Path to Enhance Large Language Models},
  author    = {Ling, Guoming and Huang, Zhongzhan and Lin, Yupei and Li, Junxin and Zhong, Shanshan and Wu, Hefeng and Lin, Liang},
  booktitle = {Findings of the Association for Computational Linguistics: ACL 2026},
  pages     = {22900--22933},
  year      = {2026}
}

@article{qin2024idam,
  title   = {An Introspective Data Augmentation Method for Training Math Word Problem Solvers},
  author  = {Qin, Jinghui and Huang, Zhongzhan and Zeng, Ying and Zhang, Quanshi and Lin, Liang},
  journal = {IEEE/ACM Transactions on Audio, Speech, and Language Processing},
  volume  = {32},
  pages   = {3113--3127},
  year    = {2024}
}

@inproceedings{huang2025minilongbench,
  title     = {{MiniLongBench}: The Low-cost Long Context Understanding Benchmark for Large Language Models},
  author    = {Huang, Zhongzhan and Ling, Guoming and Zhong, Shanshan and Wu, Hefeng and Lin, Liang},
  booktitle = {Proceedings of the 63rd Annual Meeting of the Association for Computational Linguistics (Volume 1: Long Papers)},
  pages     = {11442--11460},
  year      = {2025}
}

@inproceedings{zhong2024clot,
  title     = {Let's Think Outside the Box: Exploring Leap-of-Thought in Large Language Models with Creative Humor Generation},
  author    = {Zhong, Shanshan and Huang, Zhongzhan and Gao, Shanghua and Wen, Wushao and Lin, Liang and Zitnik, Marinka and Zhou, Pan},
  booktitle = {Proceedings of the IEEE/CVF Conference on Computer Vision and Pattern Recognition},
  pages     = {13246--13257},
  year      = {2024}
}

@inproceedings{huang2021rethinking,
  title     = {Rethinking the Pruning Criteria for Convolutional Neural Network},
  author    = {Huang, Zhongzhan and Shao, Wenqi and Wang, Xinjiang and Lin, Liang and Luo, Ping},
  booktitle = {Advances in Neural Information Processing Systems},
  volume    = {34},
  pages     = {16305--16318},
  year      = {2021}
}

@inproceedings{he2021blending,
  title     = {Blending Pruning Criteria for Convolutional Neural Networks},
  author    = {He, Wei and Huang, Zhongzhan and Liang, Mingfu and Liang, Senwei and Yang, Haizhao},
  booktitle = {Artificial Neural Networks and Machine Learning -- ICANN 2021},
  series    = {Lecture Notes in Computer Science},
  pages     = {3--15},
  publisher = {Springer},
  year      = {2021}
}

@article{ling2026bestrand,
  title   = {Random Is Almost Enough? Stable and Low-Cost Benchmark Compression via {BestRand}},
  author  = {Ling, George and Zhong, Shanshan and Yang, Lijin and Yang, Hao and Li, Nanxiang and Huang, Zhongzhan},
  journal = {arXiv preprint arXiv:2611.00883},
  year    = {2026}
}

@article{ling2026survey,
  title   = {Are We Testing Too Much? A Survey of Benchmark Compression for Efficient Evaluation of Large Language Models},
  author  = {Ling, Guoming and Lin, Yupei and Huang, Zhongzhan and Li, Junxin and Liu, Junjie and Zhong, Shanshan and Wu, Hefeng and Lin, Liang},
  journal = {arXiv preprint arXiv:2610.18248},
  year    = {2026}
}

@article{fan2026language,
  title={Language Models Can Rewrite Their Own Weights},
  author={Fan, Jiajun and He, Neil and Yang, Changyi and Mao, Andy and Zhu, Xi and Cheng, Chaoran and Liang, Kaiqu and Shen, Shuaike and Li, Zetian and Guo, Ruihan and others},
  journal={arXiv preprint arXiv:2611.21885},
  year={2026}
}

@article{kundu2026same,
  title={Same Fix, Different Weights: Replicating Weight-Space Programming with Claude},
  author={Kundu, Subham},
  journal={arXiv preprint arXiv:2611.20953},
  year={2026}
}

@article{hu2026exploring,
  title={Exploring weight-space programming: Capabilities, generalization, and limitations},
  author={Hu, Junru and Tang, Cheng and Li, Wei and Zhu, Hongze and Feng, Chun-Mei and Zhang, Qi and Chen, Yirong and Hu, Ming},
  journal={arXiv preprint arXiv:2611.09382},
  year={2026}
}

@article{fischetti2026just,
  title={Just sort it! A note on the LP behind rank-one anti-loop edits},
  author={Fischetti, Matteo},
  journal={arXiv preprint arXiv:2611.21432},
  year={2026}
}

@article{salloum2026tree,
  title={Tree-Ising: Repairing Language Models with Bit-Level Ising Edits},
  author={Salloum, Hadi and Ali, Ammar},
  journal={arXiv preprint arXiv:2611.14210},
  year={2026}
}

\appendix

\section{The Editing Interface}
\label{app:interface}

\begin{figure*}[t]
\begin{tcolorbox}[colback=black!2,colframe=black!22,boxrule=0.4pt,arc=1.5pt,left=6pt,right=6pt,top=4pt,bottom=4pt,
  title={\small\bfseries System prompt},coltitle=black,colbacktitle=black!8,toptitle=2pt,bottomtitle=2pt]
\scriptsize\ttfamily\raggedright\setlength{\parskip}{3pt}
You are Qwen3.5-0.8B. Improve your own numerical weights for mathematics, without gradients or training. The target always solves questions with thinking enabled. Practice feedback is optimization data; tests are hidden. First choose one decoder layer, then one or two tensors of that layer to read. You will receive statistics, sampled real values and row/column norms of the current weights. Then propose an explicit edit. The host stops at the fixed proposal budget or patience limit; do not stop yourself. Positions and values must be your own decisions; no successful repair is provided. Repairing an internal failure is allowed; do not make the target skip thinking or close it immediately regardless of the question.

Allowed layouts: flat entries, whole rows, or whole columns. start/count select consecutive indices with circular wrap. Operations: zero; scale by value; add value; fill with value; normal replacement with this tensor mean and abs(value) times its standard deviation; project subtracts value times the component along a normalized column direction you choose; expression computes the replacement with bounded NumPy. For expression, S is the selected slice, X the current full tensor, B its original, T0/T1 the inspected tensors. Supported expressions include array slicing, + - * / @, np.mean/std/sum/outer/dot/array/reshape and np.linalg.norm. The expression must return a scalar or the same shape as S.

No files, internet, optimizer, architecture changes or whole-tensor uniform rescaling. At most 8 million BF16 values may differ from the original model. An edit is adopted only if neither of two disjoint practice batches gets worse and\newline together they gain at least 3 correct answers. Adopted weights generate your next decision. Give a brief hypothesis, not a claim of proven mechanism.

Respond using the supplied JSON schema; do not write a Python script or invent extra tools. This interface requires stop=false when selecting tensors and action=edit after reading them. These are task requirements, not evidence that an edit is beneficial. Choose every tensor, location, count, operation and numerical value yourself. Invalid or ineffective proposals remain failures; the host never repairs their semantics.

Tensor catalog grouped by decoder layer (ids are opaque labels): \{"0":[\{"id":"p273", "name":"layers.0.input\_layernorm.weight", "shape":[1024]\}, \{"id":"p298", "name":"layers.0.linear\_attn.A\_log", "shape":[16]\}, \{"id":"p313", "name":"layers.0.linear\_attn.conv1d.weight", "shape":[6144,1,4]\}, \ldots\ \{"id":"p272", "name":"layers.23.self\_attn.v\_proj.weight", "shape":[512,1024]\}]\}
\end{tcolorbox}
\caption{The editor's system prompt under the ordered interface, verbatim from the run of Figure~\ref{fig:example}, with only the first and last entries of the 318-entry tensor catalog kept. The prompt of the shuffled interface differs in three places. It asks the editor to pick one or two tensors directly, states the keep rule as an improvement in the combined accuracy of the two batches, and gives the catalog as a shuffled list without layers.}
\label{fig:prompt}
\end{figure*}

\begin{table*}[t]
\centering
\footnotesize
\setlength{\tabcolsep}{6pt}
\renewcommand{\arraystretch}{1.25}
\newcommand{\js}[1]{{\scriptsize\texttt{#1}}}
\begin{tabular}{@{}l>{\raggedright}p{3.9cm}ll@{}}
\toprule
Call & What the editor newly sees & Required JSON output & Constraint \\
\midrule
1\enspace Layer & System prompt, tensor catalog and practice feedback & \js{\{"stop": false, "layer": $L$\}} & $L$ from 0 to 23 \\
2\enspace Tensors & All tensors of the chosen layer & \js{\{"tensors": [...]\}} & One or two identifiers \\
3\enspace Edit & Statistics, sampled values and row and column norms of the one or two chosen tensors & \js{\{"action": "edit", "edits": [...], "hypothesis": "..."\}} & One or two edits \\
\bottomrule
\end{tabular}
\caption{The three calls of one proposal under the ordered interface. The shuffled interface has no first call, and the editor picks tensors directly from the shuffled catalog.}
\label{tab:calls}
\end{table*}

The editor works with thinking turned off. Table~\ref{tab:calls} lists the three calls of one proposal under the ordered interface, and the shuffled interface needs only two because it has no layer step. All calls share the system prompt and the current practice feedback, and each call sees the questions and answers of the previous ones. Figure~\ref{fig:prompt} gives the system prompt verbatim.

\paragraph{Tensor catalog and identifiers.}
The catalog gives every editable tensor a random identifier such as \code{p298}. The identifier carries no information, and the same tensor has different identifiers in different runs. The catalog also lists the full name and shape behind each identifier, for example \code{\{"id": "p298", "name": "layers.0.linear\_attn.A\_log", "shape": [16]\}}. In later calls the editor refers to tensors only by identifier, and the grammar accepts only identifiers that appear in the catalog. In Figure~\ref{fig:example}, \code{p273} and \code{p298} are the identifiers of the layer-0 input norm and \alog in that run. The ordered interface groups the catalog by layer and sorts each layer by name, while the shuffled interface lists all tensors in a random order without layers. Table~\ref{tab:tensors} explains the tensors that appear in this paper.

\begin{table}[t]
\centering
\footnotesize
\setlength{\tabcolsep}{5pt}
\renewcommand{\arraystretch}{1.12}
\renewcommand{\code}[1]{{\scriptsize\texttt{#1}}}
\begin{tabular}{@{}ll@{}}
\toprule
Tensor & Role \\
\midrule
\grpn{2}{Linear-attention layers}
\code{A\_log} & How fast each head forgets \\
\code{dt\_bias} & Per-head bias of the forgetting rate \\
\code{in\_proj\_a} & Input-dependent forgetting rate \\
\code{in\_proj\_b} & Strength of writes to memory \\
\code{in\_proj\_z} & Output gate \\
\code{conv1d} & Short convolution before attention \\
\code{linear\_attn.norm} & Output normalization \\
\code{out\_proj} & Output projection \\
\grpn{2}{Standard-attention layers}
\code{q\_norm} & Query normalization \\
\grpn{2}{All layers}
\code{input\_layernorm} & Normalization before attention \\
\code{post\_attention\_layernorm} & Normalization before the MLP \\
\code{gate\_proj} & Gate matrix of the MLP \\
\code{up\_proj} & Up-projection of the MLP \\
\code{down\_proj} & Down-projection of the MLP \\
\bottomrule
\end{tabular}
\caption{Tensors that appear in this paper and their roles. \code{A\_log} holds one number per head, 16 per layer.}
\label{tab:tensors}
\end{table}

\paragraph{Practice feedback.}
For the first practice batch, the practice feedback reports the current model's correct answers, accuracy, share of unfinished answers, mean response length and degree of repetition, together with the question, reference answer, prediction and last 600 characters of the response for its first four questions. It also gives summary scores on the second batch and the edits and outcomes of recent proposals. Each call takes 11.6k to 23k input tokens and returns 15 to 150 tokens.

\paragraph{Edit format and rejection rules.}
An edit names a tensor identifier, a layout, a start, a count and an operation. The layout can be flattened entries, whole rows or whole columns, and positions are taken consecutively from the start, wrapping around at the end. Operation values must lie between $-10$ and $10$. The following edits are rejected outright, counted as invalid and not retried: rescaling a whole tensor uniformly, repeating an evaluated edit, changing no value after rounding to BF16, altering more than 8 million values, and exceeding the context length. Embeddings, the output layer and the final normalization cannot be edited.

\section{Data, Search and Evaluation}
\label{app:protocol}

\begin{table}[t]
\centering
\footnotesize
\setlength{\tabcolsep}{5pt}
\renewcommand{\arraystretch}{1.12}
\begin{tabular}{@{}lrr@{}}
\toprule
 & GSM8K & MATH \\
\midrule
\grpn{3}{Training split}
Official training split & 7,473 & 7,500 \\
Removed questions & 2,378 & 227 \\
Usable training questions & 5,095 & 7,273 \\
\quad Practice pool & 3,559 & 5,737 \\
\quad Held-out questions & 1,536 & 1,536 \\
\quad\quad Validation, shuffled & 256 & 256 \\
\quad\quad Validation, ordered & 512 & 512 \\
\quad\quad Random-edit screening & 128 & 128 \\
\grpn{3}{Test split}
Test questions & 1,319 & 500 \\
\bottomrule
\end{tabular}
\caption{Question splits of each dataset. Indented rows are carved out of the row above. Validation and screening questions come from the held-out questions and are therefore part of the usable training questions.}
\label{tab:data}
\end{table}

\paragraph{Data splits.}
Table~\ref{tab:data} gives the question splits. GSM8K and MATH are searched separately, each using only its own training questions. The official MATH training split has 7,500 questions. We reuse the practice set released with the GPT-6 experiments \citep{takehi2026weightspace}, which keeps 7,275 of them and removes 225 that lack an answer or share a span of 13 consecutive words with an evaluation question. We further drop 2 questions whose text exactly matches a test question, leaving 7,273. The official GSM8K training split has 7,473 questions, of which 2,378 had already been used in experiments before this study and are removed, leaving 5,095. The usable training questions are then split into a practice pool and 1,536 held-out questions that do not overlap, and the validation and screening questions are carved out of the held-out questions. Validation questions are therefore part of the usable training questions rather than additional to them. Every round of every run draws two fresh, non-overlapping practice batches from this pool.

\begin{table}[t]
\centering
\footnotesize
\setlength{\tabcolsep}{4pt}
\renewcommand{\arraystretch}{1.12}
\begin{tabular}{@{}lcc@{}}
\toprule
 & Shuffled & Ordered \\
\midrule
\grpn{3}{Catalog and calls}
Catalog order & random & by layer \\
Editor picks & tensors & layer, tensors \\
Calls per proposal & 2 & 3 \\
\grpn{3}{Practice and validation}
Practice questions per batch & 32 & 128 \\
Gain needed to keep an edit & 1 & 3 \\
Validation questions & 256 & 512 \\
\grpn{3}{Budget and scale}
Valid edits per round & 4, 6, 12 & 8 \\
Most proposals per round & 24 & 24 \\
Rejections in a row to end & 8 & 6 \\
Rounds per run & 3 & 4 or 8 \\
Runs per setting & 24 & 24 or 12 \\
\bottomrule
\end{tabular}
\caption{Settings of the two interfaces. The gain needed to keep an edit is the total over both practice batches. The shuffled interface spends 4, 6 and 12 valid edits in rounds 1 to 3. Under the ordered interface, the setting from the original model and the +4 setting each have 24 runs of 4 rounds, and the eight-round continued settings each have 12 runs.}
\label{tab:settings}
\end{table}

\paragraph{Search settings.}
Table~\ref{tab:settings} gives the settings of the two interfaces. The GSM8K +8 runs start from the three selected shuffled models A, B and C, the MATH +4 runs from the six selected shuffled models, and the MATH +12 runs from the three selected +4 models. Every starting model has 4 runs.

\paragraph{Selection and evaluation.}
For each run we take as its best model the round with the largest validation gain among the states that have kept an edit and differ from the starting point. For each setting we then take the three runs with the largest such gain as the selected models. Their weights are frozen and recorded before testing. We evaluate all 1,319 GSM8K test questions with two samples each and all 500 MATH500 questions with four samples each, with thinking enabled, a 32,768-token limit, temperature 0.6, top-p 0.95, top-k 20 and presence penalty 1.5, and report mean accuracy. Each edited model is compared with the original model run on the same hardware and sampling seeds by a paired bootstrap over questions with Bonferroni correction. All spot-check runs are fixed before the search and tested whatever their outcome, and they are of two kinds. The first kind comprises six runs per dataset in the ordered setting from the original model, chosen by a fixed hashing rule and tested after their last round. The second kind is the first run of each starting model in the eight-round continued settings, tested after its 4th and 8th rounds.

\section{Random Baselines}
\label{app:random}

\begin{table}[t]
\centering
\footnotesize
\setlength{\tabcolsep}{5pt}
\renewcommand{\arraystretch}{1.12}
\begin{tabular}{@{}l>{\raggedright\arraybackslash}p{5.25cm}@{}}
\toprule
 & Sampling \\
\midrule
\grpn{2}{Random search}
Tensor & One drawn uniformly,\newline or two with probability 0.2 \\
Layout, start & Uniform, with flat entries, rows or columns for matrices \\
Count & 1 to 131,072, half at fixed levels\newline and half log-uniform \\
Operation & Zero, scale, redraw, add or fill a number, remove a direction \\
Scale factors & 0, 0.2, 0.5, 1, 2, 3, $-1$ \\
Selection & Same as shuffled-interface self-editing \\
\grpn{2}{Single random edits}
Tensor, layout & As above, but one tensor only \\
Count & 10 levels 1, 4, 16 up to 131,072, uniform \\
Operation & Zero, nine scales and redraw, 32 each \\
Scale factors & $\pm0.2$, $\pm0.5$, $-1$, $\pm2$, $\pm3$ \\
Selection & None, applied once to the original model \\
\bottomrule
\end{tabular}
\caption{Sampling of the two random baselines.}
\label{tab:random}
\end{table}

All random edits are drawn without any model call and applied by the same code that applies the editor's edits. Table~\ref{tab:random} summarizes how the two random baselines are sampled.

\paragraph{Random search.}
The redraw scale $|v|$ is 0.2, 0.5, 1, 2 or 3. Added or filled numbers are $\pm10^{u}$ with $u$ uniform between $-3$ and 0. Removing a direction applies only to the column layout, with a coefficient of 0.2, 0.5, 1, 1.5, 2 or 3. With the same budget, random search spends exactly as many valid edits per round as the self-editing run with the same random seed. The larger-budget settings spend a fixed 4, 6 or 12 valid edits per round, in total about twice the same budget on GSM8K and 1.7 times on MATH. The larger-budget GSM8K setting also extends the scale factors to 0, $\pm0.2$, $\pm0.5$, $\pm1$, $\pm2$ and $\pm3$. On MATH we ran two larger-budget settings, one with unchanged and one with extended scale factors. The main text and Figure~\ref{fig:random_search} report the former, and the best model of the latter reaches only 7.5\%.

\paragraph{Single random edits.}
We draw 32 edits for each of eleven operations, namely setting to zero, nine scales and redrawing from the tensor's mean and standard deviation. Each edit goes to a random tensor, layout, start and count and is applied only once to the original model. The screening samples each of 128 held-out training questions once with an 8,192-token limit. Each screening task also lets the original model answer again with another random seed, which gives 45 or 46 score changes, and the grey band of Figure~\ref{fig:random_edits} spans their 5th to 95th percentile. All selected models of the shuffled interface, the selected MATH500 +4 models and GPT-6's two edits take part in the same screening for reference. The best eight random edits per dataset are rescored on 512 further questions from the practice pool, and the best of them is tested.

\section{The Edits Behind the Selected Models}
\label{app:edits}

Tables~\ref{tab:edits}, \ref{tab:edits_math} and~\ref{tab:edits_math2} list in order every kept edit behind all 24 selected models, transcribed from the editor's JSON output, and Table~\ref{tab:tensors} explains the tensors. All 15 models from the ordered interface edit \alog in layer 0, while the models from the shuffled interface edit a wide range of tensors.

\begin{table*}[p]
\centering
\footnotesize
\setlength{\tabcolsep}{3.5pt}
\renewcommand{\arraystretch}{1.0}
\renewcommand{\code}[1]{\texttt{#1}}
\begin{tabular}{@{}l c l l l >{\raggedright\arraybackslash}p{4.05cm}@{}}
\toprule
Proposal & Layer & Tensor & Block & Operation & Note \\
\midrule
\grp{Shuffled interface, from the original model}
\mdl{Top-1, model A}{test accuracy 31.9\%, 8,192 values changed}
g1\#1 & 12 & \code{in\_proj\_b} & rows 12--15, 0--3 & scale $-0.5$ & Hypothesis says first 8 entries \\
\mdl{Top-2, model B}{test accuracy 28.7\%, 1,048,686 values changed}
g2\#0, g2\#4 & 4 & \code{linear\_attn.norm} & [10:14] & redraw, $|v|=0.168$ & Both edits in both proposals \\
 & 7 & \code{post\_attention\_layernorm} & [10:14] & redraw, $|v|=0.168$ & \\
g3\#0 & 14 & \code{gate\_proj} & rows 159 to 1182 & scale $0.897$\cp & The two values equal the mean \\
 & 16 & \code{linear\_attn.norm} & [0:102] & add $0.087$\cp & and standard deviation it read \\
g3\#4 & 16 & \code{linear\_attn.norm} & [0:48] & scale $-0.027$ & \\
\mdl{Top-3, model C}{test accuracy 16.5\%, 16 values changed}
g1\#12 & 14 & \code{linear\_attn.norm} & [2:18] & scale $-0.07$ & \\
\addlinespace[2pt]
\grp{Ordered interface, from the original model}
\mdl{Top-1}{test accuracy 61.0\%, 48 values changed}
g1\#0 & 16 & \code{input\_layernorm} & [0:16] & scale $-0.5$ & \\
g2\#0 & 0 & \code{input\_layernorm} & [0:16] & scale $-1.979$\cp & The proposal of Figure~\ref{fig:example} \\
 & 0 & \code{A\_log} & all 16 & add $-1.979$\cp & \\
\mdl{Top-2}{test accuracy 59.9\%, 32 values changed}
g3\#2 & 0 & \code{input\_layernorm} & [0:16] & scale $-1.492$ & No hypothesis given \\
 & 0 & \code{A\_log} & all 16 & add $-3.031$ & \\
\mdl{Top-3}{test accuracy 54.0\%, 276 values changed}
g1\#0 & 0 & \code{input\_layernorm} & [0] & zero & \\
 & 0 & \code{conv1d} & [0] & scale $0.338$ & \\
g2\#1 & 0 & \code{A\_log} & [0] & scale $-2.031$ & \\
 & 0 & \code{conv1d} & [0] & add $-4.063$ & \\
g2\#11 & 18 & \code{input\_layernorm} & [0] & scale $-2.5$ & \\
 & 18 & \code{A\_log} & [0] & add $-3.75$ & \\
g4\#2 & 0 & \code{A\_log} & all 16 & add $-2.028$ & \\
g4\#3 & 17 & \code{post\_attention\_layernorm} & [492:748] & add $-1.028$ & \\
\addlinespace[2pt]
\grp{Continued +8, from selected models of the shuffled interface}
\mdl{Top-1, from model B}{test accuracy 62.2\%, 320 values changed from the start}
g2\#1 & 0 & \code{A\_log} & all 16 & add $-2.0$ & \\
g2\#6 & 20 & \code{input\_layernorm} & [0:128] & scale $-0.952$ & \\
g4\#1 & 0 & \code{in\_proj\_a} & flat [260:276] & add $-0.038$ & \\
g4\#3 & 0 & \code{in\_proj\_a} & flat [260:276] & add $-0.038$ & Repeats the edit of g4\#1 \\
 & 0 & \code{in\_proj\_b} & flat [520:536] & add $-0.038$ & \\
g6\#10 & 0 & \code{input\_layernorm} & [0:128] & scale $-0.952$ & \\
g8\#15 & 0 & \code{out\_proj} & flat [0:16] & scale $-0.550$ & \\
\mdl{Top-2, from model A}{test accuracy 53.3\%, 1,040 values changed from the start}
g1\#0 & 0 & \code{input\_layernorm} & [0:16] & add $-1.979$\cp & \\
 & 0 & \code{A\_log} & all 16 & add $-1.979$\cp & \\
g2\#4 & 0 & \code{A\_log} & all 16 & add $-0.058$ & Hypothesis swaps the values \\
 & 0 & \code{input\_layernorm} & [0:1024] & add $-0.041$ & \\
\mdl{Top-3, from model C}{test accuracy 48.1\%, 21 values changed from the start}
g2\#1 & 0 & \code{input\_layernorm} & [0] & scale $0.5$ & \\
 & 0 & \code{A\_log} & [0] & scale $0.8$ & \\
g4\#19 & 10 & \code{A\_log} & [0] & scale $-1.983$ & \\
g5\#7 & 0 & \code{A\_log} & [0:16] & scale $-1.983$ & \\
 & 0 & \code{A\_log} & all 16, start 1 & add $-3.956$ & \\
g5\#9 & 0 & \code{A\_log} & all 16, start 1 & scale $-3.956$ & \\
 & 0 & \code{A\_log} & all 16, start 2 & add $-3.956$ & \\
g6\#1 & 12 & \code{input\_layernorm} & [0:2] & scale $0.340$ & \\
g7\#1 & 12 & \code{A\_log} & [0] & scale $0.347$ & Hypothesis ``The cost of one chair is \$11.'' \\
\bottomrule
\end{tabular}
\caption{Every kept edit behind all nine selected GSM8K models. In the proposal column, g$r$\#$k$ marks proposal $k$ of round $r$ with $k$ counted from 0, and an edit with an empty proposal cell belongs to the same proposal as the row above. Slices exclude their right end, and flat counts the flattened entries of a matrix. Redrawing samples new values from a normal distribution with the tensor's mean and $|v|$ times its standard deviation. Values marked $^{\ast}$ equal a statistic the editor had just read in the same proposal. Continued runs list only the edits after their start, and their counts of changed values are relative to the start.}
\label{tab:edits}
\end{table*}

\begin{table*}[p]
\centering
\footnotesize
\setlength{\tabcolsep}{3.5pt}
\renewcommand{\arraystretch}{1.0}
\renewcommand{\code}[1]{\texttt{#1}}
\begin{tabular}{@{}l c l l l >{\raggedright\arraybackslash}p{4.05cm}@{}}
\toprule
Proposal & Layer & Tensor & Block & Operation & Note \\
\midrule
\grp{Shuffled interface, batch 1, from the original model}
\mdl{Top-1}{test accuracy 17.3\%, 16,374 values changed}
g1\#1, g1\#3 & 12 & \code{in\_proj\_b} & all 16 rows & redraw, $|v|=0.497$ & The same edit is kept twice \\
\mdl{Top-2}{test accuracy 10.4\%, 3,237 values changed}
g2\#3 & 9 & \code{out\_proj} & flat [147:1171] & scale $-0.5$ & \\
g3\#2 & 23 & \code{q\_norm} & [0:165] & zero & \\
 & 17 & \code{in\_proj\_z} & flat [0:2048] & scale $-0.004$ & \\
\mdl{Top-3}{test accuracy 8.1\%, 1 value changed}
g3\#0 & 22 & \code{input\_layernorm} & [129] & add $-4.5$ & \\
\addlinespace[2pt]
\grp{Shuffled interface, batch 2, from the original model}
\mdl{Top-1}{test accuracy 17.7\%, 1,039 values changed}
g2\#0 & 12 & \code{dt\_bias} & [2:17] & scale $-0.8$ & \\
 & 4 & \code{down\_proj} & flat [495:1519] & redraw, $v=0$ & Sets the block to the mean \\
\mdl{Top-2}{test accuracy 22.0\%, 4,639 values changed}
g2\#0 & 22 & \code{linear\_attn.norm} & [48:64] & scale $-1.035$ & \\
 & 22 & \code{down\_proj} & flat [48:64] & add $0.980$ & \\
g3\#0 & 2 & \code{in\_proj\_b} & flat [0:3584] & redraw, $|v|=0.268$ & \\
 & 2 & \code{in\_proj\_b} & flat [3584:4608] & add $-0.012$ & \\
\mdl{Top-3}{test accuracy 8.3\%, 33 values changed}
g1\#0 & 21 & \code{post\_attention\_layernorm} & [0] & zero & \\
 & 21 & \code{post\_attention\_layernorm} & [16:48] & scale $-0.5$ & \\
\addlinespace[2pt]
\grp{Continued +4, from selected models of shuffled batch 2}
\mdl{Top-1, from batch-2 Top-1}{test accuracy 25.4\%, 1,055 values changed from the start}
g3\#18 & 0 & \code{A\_log} & all 16 & add $-3.0$ & Plus one edit that scales by 1 \\
g4\#4 & 0 & \code{input\_layernorm} & [0:16] & redraw, $|v|=0.238$\cp & \\
g4\#9 & 0 & \code{post\_attention\_layernorm} & [0:1024] & redraw, $|v|=0.085$ & \\
\mdl{Top-2, from batch-2 Top-3}{test accuracy 25.4\%, 48 values changed from the start}
g2\#11 & 16 & \code{post\_attention\_layernorm} & [0:16] & zero & \\
g2\#12 & 0 & \code{input\_layernorm} & [0:16] & zero & \\
g3\#7 & 0 & \code{A\_log} & all 16 & add $-3.0$ & \\
\mdl{Top-3, from batch-2 Top-3}{test accuracy 28.4\%, 65 values changed from the start}
g1\#0 & 0 & \code{input\_layernorm} & [0:16] & scale $-1.979$\cp & \\
 & 0 & \code{A\_log} & all 16 & add $-1.979$\cp & \\
g1\#4 & 0 & \code{in\_proj\_b} & flat [0:16] & add $-1.979$ & \\
 & 0 & \code{in\_proj\_a} & flat [0:16] & scale $-1.979$ & \\
g2\#10 & 0 & \code{in\_proj\_z} & flat [0] & scale $-0.047$ & \\
 & 0 & \code{in\_proj\_a} & flat [0] & scale $-1.319$ & \\
\bottomrule
\end{tabular}
\caption{Every kept edit behind the selected MATH500 models, part one with the nine models of the shuffled interface and of continued +4, in the notation of Table~\ref{tab:edits}.}
\label{tab:edits_math}
\end{table*}

\begin{table*}[p]
\centering
\footnotesize
\setlength{\tabcolsep}{3.5pt}
\renewcommand{\arraystretch}{1.0}
\renewcommand{\code}[1]{\texttt{#1}}
\begin{tabular}{@{}l c l l l >{\raggedright\arraybackslash}p{4.6cm}@{}}
\toprule
Proposal & Layer & Tensor & Block & Operation & Note \\
\midrule
\grp{Ordered interface, from the original model}
\mdl{Top-1}{test accuracy 24.2\%, 16 values changed}
g1\#6 & 0 & \code{A\_log} & all 16 & add $-3.2$ & \\
g3\#1 & 0 & \code{A\_log} & [0:3] & scale $-5.179$ & Only these 16 values change \\
\mdl{Top-2}{test accuracy 21.9\%, 16,634 values changed}
g1\#17 & 0 & \code{input\_layernorm} & [0:245] & scale $-0.006$ & \\
 & 0 & \code{A\_log} & [3:5] & scale $-0.080$ & \\
g2\#4 & 12 & \code{A\_log} & all 16 & add $-1.492$ & Plus one edit that scales by 1 \\
g3\#5 & 0 & \code{input\_layernorm} & [0:2] & zero & \\
g4\#3 & 12 & \code{in\_proj\_a} & flat [0:16384] & fill $-0.05$ & The whole tensor \\
\mdl{Top-3}{test accuracy 24.7\%, 34 values changed}
g1\#3 & 17 & \code{A\_log} & [0:2] & scale $-1.5$ & \\
g3\#0 & 0 & \code{input\_layernorm} & [0:16] & scale $-1.979$\cp & \\
 & 0 & \code{A\_log} & all 16 & add $-1.979$\cp & \\
\addlinespace[2pt]
\grp{Continued +12, eight more rounds from selected +4 models}
\mdl{Top-1, from +4 Top-2}{test accuracy 31.8\%, 24,582 values changed from the start}
g1\#1 to g1\#3 & 0 & \code{A\_log} & [0:2] & scale $-0.973$ & Once in each of three proposals \\
g1\#2 & 0 & \code{up\_proj} & flat [0:24576] & scale $-0.973$ & \\
g2\#16 & 10 & \code{A\_log} & [0:3] & scale $-1.594$ & \\
g3\#0 & 12 & \code{input\_layernorm} & [0] & zero & \\
\mdl{Top-2, from +4 Top-2}{test accuracy 32.2\%, 2 values changed from the start}
g3\#9 & 0 & \code{A\_log} & [0] & scale $-4.979$ & Both change the same two values \\
 & 0 & \code{input\_layernorm} & [0] & add $-4.979$ & \code{A\_log}[0] $-2.742 \to 8.686$ \\
g7\#3 & 0 & \code{A\_log} & [0] & add $-4.969$ & \code{input\_layernorm}[0] $0 \to -3.578$ \\
 & 0 & \code{input\_layernorm} & [0] & add $1.395$ & \\
\mdl{Top-3, from +4 Top-1}{test accuracy 28.6\%, 2,597 values changed from the start}
g1\#0 & 10 & \code{input\_layernorm} & [0:204] & redraw, $v=0$ & Sets the block to the mean \\
 & 10 & \code{out\_proj} & flat [0:204] & redraw, $v=0$ & Sets the block to the mean \\
g5\#7 & 0 & \code{input\_layernorm} & [0:128] & scale $1.5$ & \\
 & 0 & \code{A\_log} & [0:2] & add $-0.03$ & \\
g5\#10 & 12 & \code{A\_log} & [0:12] & scale $1.5$ & \\
 & 12 & \code{in\_proj\_a} & flat [0:2048] & add $-0.03$ & \\
\bottomrule
\end{tabular}
\caption{Every kept edit behind the selected MATH500 models, part two with the six models of the ordered interface and of continued +12, in the notation of Table~\ref{tab:edits}.}
\label{tab:edits_math2}
\end{table*}

\end{document}